\pdfoutput=1

\documentclass[11pt]{article}

\usepackage{acl}

\usepackage{times}
\usepackage{float}
\usepackage{latexsym}
\usepackage{amsthm}
\usepackage{algorithm,algorithmic}
\usepackage{amsmath}
\usepackage{amssymb}
\usepackage{mathtools}
\usepackage{enumitem}
\setlist[itemize]{noitemsep, topsep=0pt, leftmargin=11pt}
\setlist[enumerate]{noitemsep, topsep=0pt, leftmargin=11pt}
\usepackage{wrapfig}
\usepackage{xcolor}
\usepackage{subcaption}
\usepackage{multirow}

\usepackage[capitalize,noabbrev]{cleveref}
\usepackage[T1]{fontenc}

\usepackage[utf8]{inputenc}

\usepackage{microtype}

\title{An Experimental Design Framework for Label-Efficient Supervised Finetuning of Large Language Models}

\author{
Gantavya Bhatt\thanks{\,\,\,Equal contribution, alphabetically ordered.} $^{,1}$, Yifang Chen$^{*,1}$, Arnav M. Das$^{*,1}$, Jifan Zhang$^{*,2}$ \\
\bf Sang T. Truong$^3$, Stephen Mussmann$^4$, Yinglun Zhu$^5$, Jeffrey Bilmes$^1$, \\
\bf Simon S. Du$^1$, Kevin Jamieson$^1$, Jordan T. Ash$^6$, Robert D. Nowak$^2$ \\
$^{1}$University of Washington, Seattle \quad
$^{2}$University of Wisconsin-Madison\\
$^{3}$Stanford University \quad
$^{4}$Georgia Institute of Technology\\
$^{5}$University of California, Riverside \quad
$^{6}$Microsoft Research NYC
}

\begin{document}
\maketitle
\begin{abstract}
Supervised finetuning (SFT) on instruction datasets has played a crucial role in achieving the remarkable zero-shot generalization capabilities observed in modern large language models (LLMs). However, the annotation efforts required to produce high quality responses for instructions are becoming prohibitively expensive, especially as the number of tasks spanned by instruction datasets continues to increase. Active learning is effective in identifying useful subsets of samples to annotate from an unlabeled pool, but its high computational cost remains a barrier to its widespread applicability in the context of LLMs. To mitigate the annotation cost of SFT and circumvent the computational bottlenecks of active learning, we propose using experimental design. Experimental design techniques select the most informative samples to label, and typically maximize some notion of uncertainty and/or diversity. In our work, we implement a framework that evaluates several existing and novel experimental design techniques and find that these methods consistently yield significant gains in label efficiency with little computational overhead. On generative tasks, to reach the same generalization performance, our methods save $50\%$ of the annotation cost compared to random sampling.
\end{abstract}

\begin{figure*}[h]
    \centering
    \includegraphics[width=.7\linewidth]{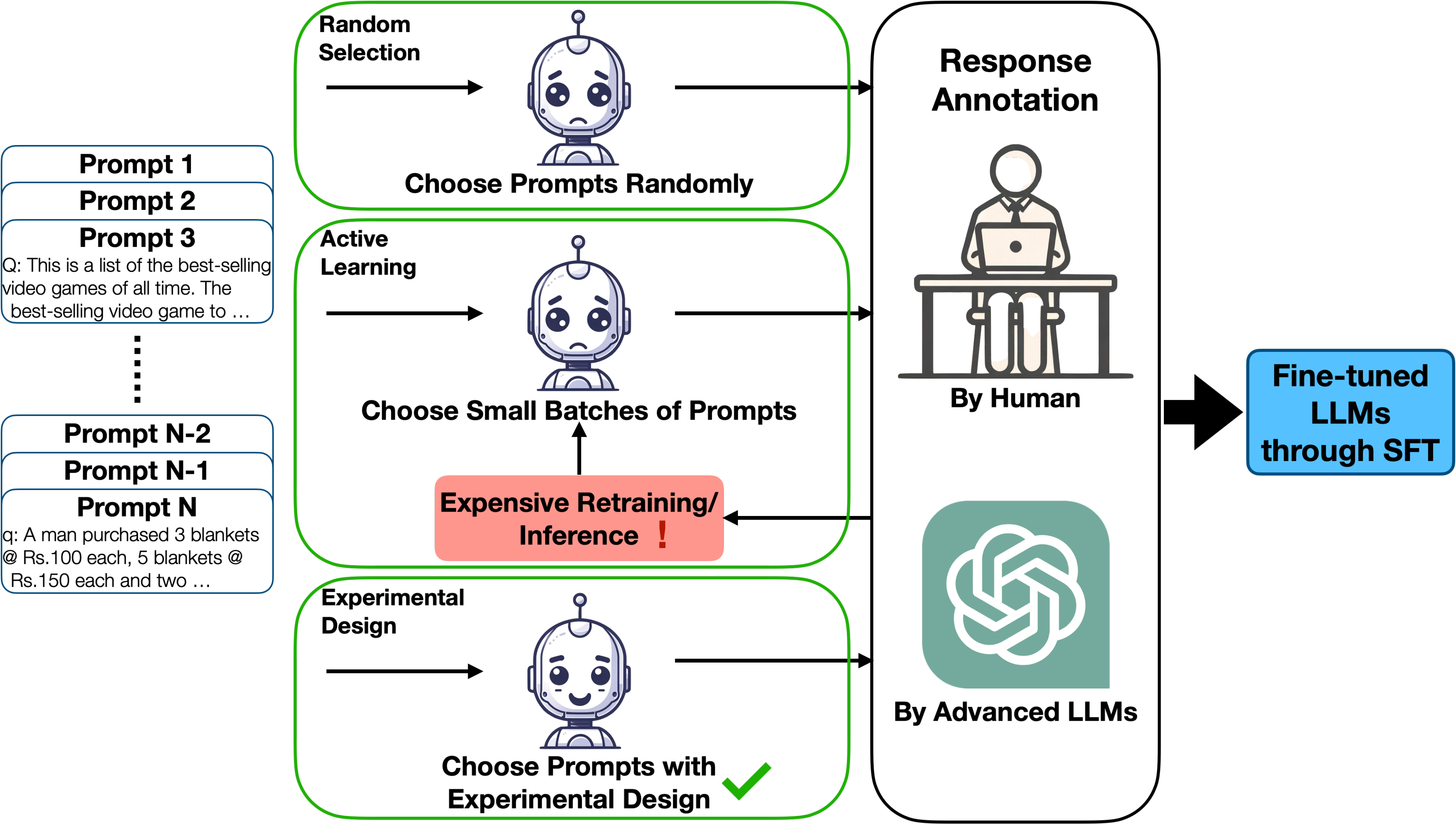}
    \caption{Comparison between different annotation schemes for label-efficient SFT. Random sampling simply chooses prompts uniformly at random which underperforms as it is prone to redundancy and may oversample from the major modes.  On the other hand, one can choose them more strategically both through active learning and experimental design. Active learning, however, is an adaptive procedure and requires computationally expensive model retraining and inference for every batch of annotation. In this paper, we study the problem through the lens of experimental design, which enjoys increased label-efficiency compared to random sampling, while incurring minimal computation cost compared to active learning.}
    \label{fig:comparison}
\end{figure*}

\section{Introduction}
Supervised finetuning (SFT) on instruction datasets has shown immense potential in improving the zero-shot performance of large language models (LLMs)~\citep{wei2022finetuned}. Recent developments in this field have been fostered by the availability of large-scale instruction datasets, consisting of natural language instructions with desired responses based on human judgment~\citep{wei2022finetuned, mishra-etal-2022-cross, longpre2023flan}. Throughout the community, there have been several efforts to further increase the number of tasks included in these datasets to improve LLM generalization~\citep{xu2022zeroprompt, wang2022supernaturalinstructions, honovich2022unnatural, wang2023selfinstruct}. In addition, supervised finetuning is especially important for handling novel forms of tasks. For example, to develop new multi-modal capabilities or defend against novel prompt hacking strategies, one must need human generated responses to finetune existing state-of-art models like GPT-4.

A crucial bottleneck of SFT is the need for annotating a massive set of instructions with detailed responses, which can be done either manually or automatically. For improving state-of-art models, one necessarily need to rely on manual approaches. This involves the use of crowd-workers or expert annotators, who produce high quality responses for almost any task but can become extremely expensive at scale due to the labor intensive process of annotation. To improve small scale and more domain-specific language models, automatic annotation methods have been proposed to reduce the burden on human annotators by labeling instructions using preexisting LLMs~\citep{honovich2022unnatural, wang2023selfinstruct, peng2023instruction}. However, models like GPT-4 are still costly to query, making the annotation for a large set of prompts potentially prohibitive.
In addition, for specialized domains, human experts are still crucial in this annotation process as general LLMs may not provide high quality responses.
Therefore, we seek to answer the following question: 

\noindent\fbox{
\parbox{.95\linewidth}{
    \emph{Can we propose label-efficient strategies that annotate fewer prompts while obtaining equally good generalization performance?}
}
}



Towards improving the label efficiency of deep models, researchers have been predominantly focusing on \emph{Active Learning}~\citep{settles2009active,gal2017deep,sener2017active,ash2019deep,zhanglabelbench}. These techniques have been proven useful in some relatively easier NLP tasks such as machine translation~\citep{honovich2022unnatural} and classification \cite{perlitz-etal-2023-active}, but remain under-explored in general natural language generation tasks. 
As shown in Figure~\ref{fig:comparison}, the active learning framework incrementally adds batches of samples to the labeled pool by repeatedly: (1) training a model on the currently labeled data and (2) using some model dependent measure of informativeness to select a new batch of points to query the annotator~\citep{atlas1989training, settles2010}. However, each iteration of active learning requires performing inference on all unlabeled samples and retraining the model on the expanded set of labeled samples~\citep{Coleman2020Selection, das2023accelerating}. In the context of parameter-heavy LLMs where inference and training are expensive, the computational cost associated with active learning may outweigh its potential savings in annotation costs. 

In this work, we propose leveraging \emph{experimental design} to select the optimal set of instructions to annotate. Experimental design concerns situations where we have to organize an experiment in order to gain some information about an object of interest. In the context of training an AI system, the “experiment” involves selecting a subset of unlabeled examples from a large pool for labeling in order to create a training set to learn a good model (the object of interest, in this case). As demonstrated in Figure~\ref{fig:comparison}, in contrast to active learning where the labeled set is expanded incrementally, experimental design techniques select the set of samples to label in \textit{a single step based solely on the initial model}. This circumvents almost all of the computational cost of active learning, allowing for gains in label efficiency to be realized with practically no overhead. While experimental design has been studied theoretically, its empirical benefits have been underexplored.

We introduce a framework for evaluating experimental design techniques for finetuning LLMs, and propose a suite of experimental design techniques that demonstrably improve the label efficiency of SFT. We develop novel scores, such as maximum token uncertainty, that quantify the LLMs uncertainty on a particular sample and correlate well with its usefulness as training data. We also propose a class of experimental design heuristics that employ the facility location function as an objective, to select a diverse and representative set of samples for annotation.




Overall, our contributions can be summarized as follows: {\bf (1)} we are the first, to the best of our knowledge, to utilize experimental design for SFT {\bf (2)} we introduce a framework to perform comprehensive evaluation on existing experimental design techniques {\bf (3)} we propose a suite of novel strategies that improve the label efficiency, significantly outperforming random sampling by more than $2\%$ accuracy across different annotation budgets {\bf (4)} compared to previous works~\citep{kung2023active,perlitz-etal-2023-active}, our work is the first to see annotation cost savings on \emph{generative} tasks. To reach the same generalization performance, our methods save $50\%$ of the annotation cost compared to random sampling (unlike random sampling that fails to achieve the same generalization).




\section{Related Work}
\paragraph{Experimental Design}

Experimental design generally refers to situations where the experiment is chosen before the collection of information (labels) starts. This is in contrast to situations where the experiment is designed in a sequential adaptive fashion, responding to information contained in labels to guide selection of the next points in the design.  This is called sequential experimental design or active learning in the parlance of machine learning.  See~\citet{pukelsheim2006optimal} for an excellent overview of classical experimental design techniques and~\citet{pronzato2013design} for a treatment of modern approaches to experimental design in nonlinear models.

Our study focuses on experimental design, rather than active learning. Active learning generally requires retraining the model and forward inference on the entire pool after each batch of labels is gathered in order to select the next batch for labeling. The retraining and repeated inference can be computationally expensive, particularly in the case of LLMs. Experimental design does not require additional computation of this sort, since the subset selection of unlabeled examples is done just once before any labels are collected.

\paragraph{Deep Active Learning} Data selection strategies for label-efficient learning have been largely studied under the framework of deep active learning, which sequentially and adaptively choose informative examples to annotate. Deep active learning methods typically use measures of uncertainty~\citep{atlas1989training, settles2010,gal2017deep,ducoffe2018adversarial,beluch2018power}, diversity~\citep{k_center_coreset,geifman2017deep,citovsky2021batch}, or some combination of both~\citep{wei2015submodularity,ash2019deep,ash2021gone,zhang2022galaxy,nuggehalli2023direct} in order to determine a set of useful samples to annotate. In addition to classical instance-level selection, there are also many existing works focused on task-wise selection \cite{xu2023cit,chen2023active,wang2023improved,fifty2021efficiently}. While we propose an experimental design framework for the label-efficient SFT problem, our strategies are inspired by the latest instance-level deep active learning literature and adapted to work under this framework for supervised finetuning of LLMs.

A few works have considered the application of active learning for SFT or other closely related settings. \citet{perlitz-etal-2023-active} explores the use of active learning to improve the label efficiency of the closely related task of natural language generation. However, this work reports inconsistent findings showing that there are limited settings where active learning demonstrates any significant advantages over random sampling. \citet{kung2023active} is the most similar to our work, proposing an active learning framework for instruction tuning. Unlike our work, \citet{kung2023active} performs task-level selection, as opposed to instance-level selection. In other words, their framework estimates the usefulness of each unlabeled task and annotates all instances within the tasks that are deemed most useful. However, task-level selection makes the simplifying assumption that every instance is equally useful within a task, which may inhibit the quality and reduce the resolution of the selected subset. Moreover, the approach in \citet{kung2023active} suffers from the aforementioned computational cost of active learning which is avoided in our framework with the use of experimental design.


\paragraph{Data Selection for SFT} Many recent works demonstrate that small subsets of instruction data can be sufficient for finetuning an LLM. \citet{zhou2023lima} demonstrate that finetuning an LLM on a dataset consisting of only manually-curated 1000 instruction/response pairs is sufficient to achieve strong generalization, but do not propose any general algorithmic procedure for subset selection. Other works propose instruction dataset pruning techniques that select subsets based on some combination of quality, diversity, and/or difficulty~\citep{chen2023alpagasus, bukharin2023data, du2023mods, li2023quantity, li2024shot}. These approaches use both the instruction and its corresponding response to choose which training samples should be retained to improve the computational efficiency of SFT. Unlike these approaches, our framework is designed to maximize label efficiency and assumes that the response to an instruction are not available until selected for annotation. Finally, \citet{hu2023validation} propose techniques to reduce the annotation cost to construct validation sets that are used for model selection, which is complementary to the method proposed in our work.

\section{Prompt Selection Strategy}
Under the experimental design framework, the learner is given a set of initial $N$ prompts $X = \{x_1, x_2, ..., x_N\}$, where each prompt $x_i \in \mathcal{X}$ is a sequence of input tokens of length $\ell_i$, $x_i = \{x_{i,1}, ..., x_{i,\ell_i}\}$, where $\mathcal{X}$ denotes the domain of all possible sequence of input tokens. Additionally, we let $g$ denote the pretrained language model. Given a budget of $k < N$, a selection strategy chooses $k$ prompts from $X$, denoted as $S \subset X$, based on different measures of informativeness of annotating an example to the pretrained model $g$. One then gathers well-written responses to prompts in $S$ from annotators (e.g., human experts or advanced LLMs). Experimental design aims to optimize the performance of the model $g'$ finetuned on the selected prompts $S$ and their responses.

\subsection{Uncertainty-Based Selection}
During the prompt selection step, since the ground truth responses are oblivious to the learner, one type of label-efficient selection strategy is to choose examples with the highest uncertainty to the model. Specifically, let $U: X \rightarrow \mathbb{R}$ define some notion of uncertainty of prompts in $X$, uncertainty-based methods simply choose the top-k most uncertain examples by
\begin{align}
    S = \operatorname*{argmax}_{\substack{S' \subset X \\ |S'| = k}} \min_{x\in S'} U(x).
\end{align}

Below, we provide four instances of uncertainty measures.
Formally, we let the pretrained model (with greedy decoding) $g: \mathcal{X} \rightarrow (\mathcal{D} \times \triangle_{\mathcal{D}})^{[L]}$ map prompts to a sequence of predicted tokens and the softmax probability distribution at each decoding step, up to length $L$. Here each token is from a dictionary $\mathcal{D}$ and each corresponding softmax probability distribution lives in the probability simplex $\triangle_{\mathcal{D}}$ over the dictionary. We let $g_y(\cdot)$ denote the sequence of $L$ tokens and $g_p(\cdot)$ denote the sequence of $L$ softmax probability distributions.


\paragraph{Mean Entropy} ~\citep{settles2010,kremer2014active}  measures the tokenwise negative entropy of the softmax probability scores. The uncertainty measure is taken as the mean across tokens as follows:
\begin{align*}
    U_{\text{entropy}}(x) = \frac{1}{|g_p(x)|}\sum_{p \in g_p(x)} \sum_{t\in \mathcal{D}} p_t \log(p_t).
\end{align*}
Where $p_t$ represents the softmax score (i.e., the probability) of the token $t$, and the entire set of $p$ values represents the distribution across the entire dictionary for a specific position in an $L$-word sentence.


\paragraph{Least Confidence}~\citep{settles2009active,settles2010} measures the model's confidence as the product of probabilities of the generated sequence. A model is more confident when the likelihood of the generated sequence is high. We take the negative confidence score as the uncertainty, which allows us to choose the least confident sequences for annotation:
\begin{align*}
    U_{\text{conf}}(x) = -\prod_{(t, p) \in g(x)} p_t.
\end{align*}


\paragraph{Mean Margin}~\citep{tong2001support,balcan2006agnostic,settles2010} measures uncertainty by taking the different between most likely and second likely token for each element in the generated sequence . A higher difference corresponds with a clearer separation between the model's best choice from its second best choice. We use the negative margin score average over all tokens as the uncertainty measure:
\begin{align*}
    \bar{U}_{\text{margin}}(x) = -\frac{1}{|g_p(x)|}\left(\sum_{p \in g_p(x)} \beta_1(p) - \beta_2(p)\right)
\end{align*}
where $\beta_1(p)$ and $\beta_2(p)$ denotes the largest and second largest element of $p$.


\paragraph{Min Margin}
is a novel strategy where we measure uncertainty based on the token with the smallest margin score instead of taking the average over all tokens. Intuitively, two sequences could have equal average token-wise margin score, but the one with smaller minimum margin is more likely to be generated as a different sequence with top-2 decoding. Mathematically, this is defined as:
\begin{align*}
    \widetilde{U}_{\text{min margin}}(x) = -\left(\min_{p \in g_p(x)} \beta_1(p) - \beta_2(p)\right).
\end{align*}

\subsection{k-Center Selection}
Another class of label-efficient selection strategy is to annotate prompts that are \emph{diverse} in the representation space.
\citet{k_center_coreset} proposed a k-center objective that chooses $k$ examples as centers of balls with equal radius. The objective promotes selections that would minimize the radius of these balls while covering all examples. Formally, the objective of choosing $k$ centers can be written as:
\begin{equation}
\label{eq:k-center}
    S  = \operatorname*{argmin}_{ \substack{ S' \subset X \\ |S'| = k} }   \operatorname*{max}_{\substack{i \in X}} \operatorname*{min}_{\substack{j \in S'}} \lVert f(x_i)-f(x_j)\rVert,
\end{equation}
where $f$ is a feature extractor mapping prompts into feature space in $\mathbb{R}^d$ and is derived from the pretrained model $h$. For decoder-only architectures, we use the last hidden state as the feature. To optimize the above NP-hard object \citep{cook1998combinatorial}, we follow the greedy methods proposed by \citet{k_center_coreset}, which enjoys a 2 multiplicative approximation guarantee to the optimal selection.

\subsection{Submodular Selection}
Equation~\eqref{eq:k-center} is commonly recognized as the minimax facility location objective \citep{springerFacilityLocation}. Additionally, we explore the conventional Facility Location (FL) function~\citep{mirchandani1990discrete}, extensively used in machine learning~\cite{wei2015submodularity, mirzasoleiman2020coresets, bilmes2022submodularity, bukharin2023data}. Given a nonnegative score $w_{ij}$ that measures the similarity between features $f(x_i)$ and $f(x_j)$, the facility location problem is formulated as follows:


\begin{equation}
\label{eq:facility_location}
    S  = \operatorname*{argmax}_{ \substack{ S' \subset X \\ |S'| = k} }   \sum_{i \in X} \operatorname*{max}_{\substack{j \in S'}} w_{ij}
\end{equation}

In Equation~\eqref{eq:facility_location}, every client $i \in X$ must have a facility within $S$, which is chosen to be the element $j \in S$ closest to $i$. FL is a known submodular function, so the greedy heuristic applied to this objective achieves a $1-1/e$ multiplicative approximation guarantee to the optimal solution~\citep{nemhauser1978analysis} despite its NP-Hard nature. The greedy algorithm can be further accelerated with the use of data structures \cite{minoux2005accelerated}, or with stochastic variants~\citep{mirzasoleiman2014lazier}.

We primarily use the radial basis function as a similarity metric, where $w_{ij} = \exp{(-\frac{\|f(x_i) - f(x_j) \|^2}{\gamma})}$ and $\gamma > 0$ is a controllable hyperparameter often referred to as the kernel width. Intuitively, $\gamma$ tunes is the degree of similarity between two data points. As $\gamma$ decreases, the similarity between $x_i$ and $x_j$ also decreases. In limiting case as $\gamma \to 0$, similarity $w_{ij} = 0$ for all $i \neq j$. For situations where hyperparameter tuning is not feasible, we propose an alternate function where the similarity metric is fixed and not tuned, where $w_{ij} = \max\{0, \frac{f(x_i)^T f(x_j)}{\|f(x_i) \| \| f(x_j) \|}\}$.




\begin{table*}[ht!]
    \centering
    \begin{tabular}{lccccc}
        \hline
        \shortstack{Strategy} & $k=0$ & $k=20K$ & $k=30K$ & $k=45K$\\
        \hline
        Random & $37.58$ & $44.33 (\pm 0.32)$ & $44.91 (\pm 0.50)$ & $45.99 (\pm 0.37)$ \\
        Mean Entropy & $37.58$ & $43.85 (\pm 0.14)$ & \underline{$45.38$}$(\pm0.21)$ & \underline{$46.45$}$(\pm 0.34)$\\
        Confidence & $37.58$ & $43.26 (\pm 0.58)$ & $44.56 (\pm 0.42)$ & \underline{$46.55$}$(\pm 0.25)$ \\
        Mean Margin & $37.58$ & $43.85 (\pm 0.33)$ & $44.88 (\pm 0.23)$ & \underline{$46.40$}$(\pm 0.16)$ \\
        Min Margin & $37.58$ & \underline{$44.55$}$(\pm 0.32)$ & \underline{$45.62$}$(\pm 0.14)$ & $45.31 (\pm 0.14)$ \\
        k-Center & $37.58$ & $43.77 (\pm 0.47)$ & \underline{$46.14$}$(\pm 0.12)$ & \underline{$46.27$}$(\pm 0.14)$ \\
        FL ($\operatorname{cosine}$) & $37.58$ & $43.77 (\pm0.23)$ & $\underline{45.89} (\pm0.50)$  & $\underline{47.01} (\pm0.37)$  \\
        FL ($\gamma=.002$) & $37.58$ & \underline{$\textbf{45.08}$}$(\pm 0.33)$ & \underline{$\textbf{47.12}$}$(\pm 0.35)$ & \underline{$\textbf{47.63}$}$(\pm 0.24)$ \\
        \hline
    \end{tabular}
    \caption{Massive Multitask Language Understanding (MMLU) evaluation of models trained on subsets selected by strategies from a pool of 99k under different annotation budgets. Each result of random strategy is averaged over 6 seeds due to the high variance from both data selection and training. Other results are averaged over 3 random seeds where the randomness mainly comes from the training. The confidence intervals are based on standard error.}
    \label{tab:mmlu}
\end{table*}

\begin{table*}[ht!]
    \centering
    \begin{tabular}{lccccc}
        \hline
        \shortstack{Strategy} & $k=0$ & $k=20K$ & $k=30K$ & $k=45K$\\
        \hline
        Random & $37.66$ & $38.95 (\pm 0.48)$ & $39.42 (\pm 0.56)$ & $39.44 (\pm 0.52)$ \\
        Mean Entropy & $37.66$ & $\underline{40.28} (\pm 0.60)$ & $38.18$$(\pm 0.43)$ & \underline{$39.99$} $(\pm 0.67)$\\
        Confidence & $37.66$ & $\underline{\textbf{40.33}} (\pm 0.49)$ & $38.28 (\pm 0.39)$ & $\underline{41.04} (\pm 0.74)$ \\
        Mean Margin & $37.66$ & $38.33 (\pm 0.67)$ & $\underline{40.05} (\pm 0.39)$ & $38.43 (\pm 0.34)$ \\
        Min Margin & $37.66$ & $\underline{39.74}$ $(\pm 0.19)$ & $\underline{40.20}$ $(\pm 0.20)$ & \underline{$39.66$} $(\pm 0.37)$ \\
        k-Center & $37.66$ & $37.44 (\pm 0.60)$ & $38.35$ $(\pm 0.63)$ & $38.6$ $(\pm 0.39)$ \\
        FL ($\operatorname{cosine}$) & $37.66$ & $38.25 (\pm0.35)$ & $\underline{39.82} (\pm0.25)$ & $\underline{40.46} (\pm1.03)$ \\
        FL ($\gamma=.002$) & $37.66$ & $38.33$ $(\pm 0.27)$ & \underline{$\textbf{41.12}$}$(\pm 0.71)$ & \underline{$\textbf{41.30}$}$(\pm 0.60)$ \\
        \hline
    \end{tabular}
    \caption{Big-Bench-Hard chain-of-thoughts (BBH-CoT) evaluation of models trained on subsets selected by strategies from a pool of 99k under different annotation budgets. Each result of random strategy is averaged over 6 seeds where each result of other strategies is averaged over 3 random seeds with standard error shown as confidence interval.}
    \label{tab:bbh}
    \vspace{-\intextsep}
\end{table*}

\section{Experiments}
In this section, we compare multiple experimental design strategies against random sampling and observe significant improvements. We describe the various setups in Section~\ref{sec:setup} and report the evaluation results based on common benchmarks~(Section~\ref{sec:eval}) and comparisons by GPT-4~(Section~\ref{sec:alpaca}). Lastly, we conduct ablation study and document the selection of hyperparameters in Section~\ref{sec:hyperparam}. Details about our experimental compute cost and complexity can be found in Appendix~\ref{sec:compute}.

\subsection{Experiment Setup} \label{sec:setup}

\paragraph{Dataset} FLAN V2~\citep{longpre2023flan} is a widely-used instruction fine-tuning dataset that combines FLAN 2021, P3++, Super-Natural Instructions, along with additional reasoning, dialogue, and program synthesis datasets. We utilize a 99K subset of FLAN V2, processed by~\citep{wang2023far}, as our training data pool, from which we select prompts and annotations (i.e., responses). 

\paragraph{Models and Training Procedure} We conduct experiments with the 7B version of the prefix language model LLaMA-2~\citep{touvron2023llama}, across different annotation budgets. Prior to fine-tuning, we choose a subset of prompts for annotation by either random sampling or using experimental design strategies, including uncertainty-based selection, k-Center selection, and submodular selections. These strategies are computed based on the prefix model only. Subsequently, we fine-tune the model on the annotated prompt/response pairs using Low-rank Adaptation (LoRA)~\citep{hu2021lora}. 

\paragraph{Evaluation metrics} 
We adopt the similar evaluation tasks as in the original FLAN V2, using the MMLU~\citep{hendrycks2020measuring} and BBH~\citep{suzgun2022challenging} benchmarks to evaluate the zero-shot generalization capability of our fine-tuned model. Massive Multitask Language Understanding dataset (MMLU) is a classification task with a set of questions about 57 subjects ranging in difficulty from elementary levels to professional levels, broadly testing mode's factual knowledge and reasoning. Big-Bench-Hard (BBH) is a generation task with 23 challenging tasks from Big-Bench~\citep{srivastava2022beyond}, broadly testing models’ general reasoning capabilities. In our study, we specifically select 5-shot MMLU and a random 20\% subset of Chain-Of-Thought BBH inputs, due to the computational resource limitations. In addition, we also evaluate our methods based on AlpacaEval~\citep{alpaca_eval}, where given a prompt, the responses from two models are compared by GPT-4 turbo. The performance of the two models are reported as the win rate against each other.

\subsection{Evaluation on Standard Benchmarks} \label{sec:eval}
When comparing different experimental design (i.e. prompt selection) strategies to random sampling in Tables~\ref{tab:mmlu},\ref{tab:bbh}, we see almost dominant improvements (around 1\% to 2\%) on Facility Location ($\gamma$ = 0.002) strategy except on the 20K BBH-CoT case (where it closely aligns with that of random sampling). Compared to 90K budget where we get MMLU = $47.76 (\pm 0.57)$, BBH-CoT = $40.49(\pm 0.30)$, we save approximately 50\% annotation cost with respect to both classification (MMLU) and generation (BBH) tasks. The facility location function with the cosine kernel also exhibits improvement over random sampling, although the magnitude of improvement is not as pronounced as observed with the tuned kernel. This disparity could be attributed to the saturation of the Facility Location function, wherein the greedy order fails to generate a diverse summary conducive to downstream applications. In practice, since one could only choose one set of examples to annotate, we provide details in how to choose the hyperparameter in Section~\ref{sec:hyperparam} before annotation and demonstrate the robustness through an ablation study. In addition, our proposed Min Margin strategy also gains much larger improvements compared to others, including the commonly used Mean Margin score.


On the other hand, we observe that most uncertainty-based selection approaches did not improve over random sampling. In some cases, they did not even exhibit monotonic improvements with the selection size. This phenomenon has a well-known explanation -- uncertainty-based methods often annotate similar/redundant examples, which could potentially hinder generalization performance when fine-tuning the model.

\begin{figure*}[t!]
    \centering
    \includegraphics[width=.9\linewidth]{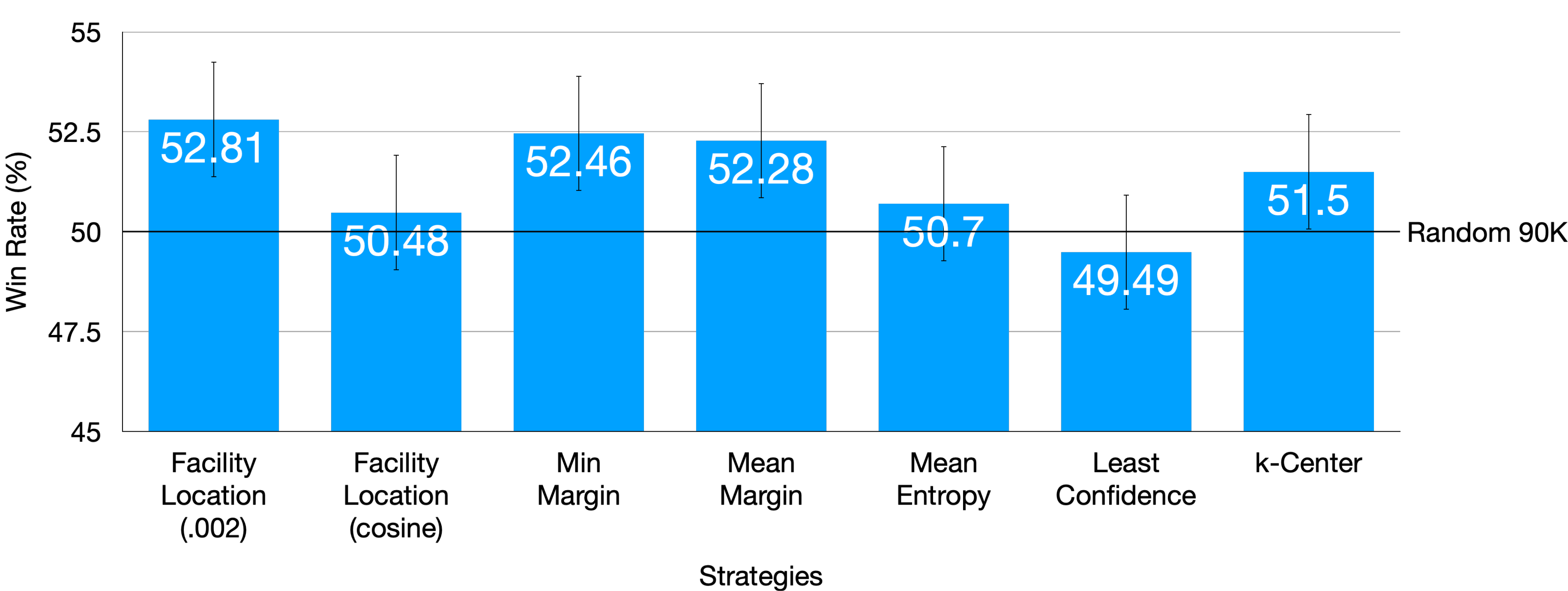}
    \caption{Evaluation by GPT-4 turbo by comparing model trained on 45K prompts selected by various strategies with the model trained on 90K random prompts. We use the win rate weighted by the continuous preferences of the GPT-4 turbo model. Error bars are reported as the standard errors across prompts.}
    \label{fig:alpaca}
    \vspace{-\intextsep}
\end{figure*}

\paragraph{Ablation} We now consider a combination of uncertainty-based selection with diversity, in the following \cref{eq: mixture}. 

\begin{equation}
\label{eq: mixture}
\begin{split}
    S  = \operatorname*{argmax}_{ \substack{ S' \subset X \\ |S'| = k} } & \sum_{i \in X}  \operatorname*{max}_{\substack{j \in S'}} w_{ij} \\  +&  \operatorname{log}\left( 1 + \sum_{x \in S'} \widetilde{U}_{\text{min margin}}(x) \right) 
\end{split}
\end{equation}

\begin{table}[]
\begin{tabular}{lcc}
\hline
                & BBH-CoT   & MMLU  \\ \hline
FL+MinMargin & 40.71($\pm$0.63) & 46.26($\pm$0.36)\\ 
MinMargin & 39.66($\pm$0.67) & 45.31($\pm$0.14)\\ 
FL  & 41.30($\pm$0.60) & 47.63($\pm$0.24)\\ 
\hline
\end{tabular}
\caption{Big-Bench-Hard chain-of-thoughts (BBH-CoT) and Massive Multitask Language Understanding (MMLU) evaluation of mixture of diversity and uncertainty strategy at the budget of 45K, averaged over 3 random seeds with standard error shown as confidence interval. We observe that the performance of the mixture interpolated between the two methods.}
\label{tab: mixture}
\vspace{-\intextsep}
\end{table}

The first part governs the diversity and the latter part governs the uncertainty, for which, we consider the min-margin as the uncertainty score. Note that the objective in \cref{eq: mixture} is a submodular maximization and hence we can use the greedy algorithm. This is a corollary of the fact that modular functions ($\sum_{x \in S'} \Tilde{U}(x)$) when composed with monotone, non-decreasing concave function is submodular, and submodularity is closed under conic combination \cite{bilmes2017deep, lin2011class, bhatt2024deep}. In general, one can apply any monotone non-decreasing concave function, which becomes a design choice. Here, we consider $\operatorname{log}(1 + x)$ which was chosen by \citet{das2023accelerating}. We consider min-margin for the uncertainty and RBF kernel based facility location. \cref{tab: mixture} shows the BBH-CoT and MMLU metrics for the mixture of min-margin and diversity at a budget of 45K and observe the performance of the mixture interpolated between the two methods. We posit that with the right choice of uncertainty metric, it is possible to improve beyond solely using uncertainty and diversity when mixed appropriately. 


\begin{table*}[t!]
    \centering
    \begin{tabular}{lccccc}
    \hline
    & Random & $\gamma=0.001$  & $\gamma=0.002$ & $\gamma=0.003$  & $\gamma=0.004$ \\ \hline
    MMLU &  $45.99 (\pm 0.37)$  &    $47.04 (\pm 0.69)$              &          $47.63 (\pm 0.24)$       &    $47.96 (\pm 0.25)$              &      $47.94 (\pm 0.49)$            \\
    BBH-CoT & $39.44 (\pm 0.52)$ & $40.56 (\pm 0.86)$ & $41.30 (\pm 0.60)$ & $41.99 (\pm 0.73)$ & $41.35 (\pm 1.04)$ \\ \hline
    \end{tabular}
    \caption{Sensitivity to kernel width $\gamma$ at 45K budget. Each result of random strategy is averaged over 6 seeds where each result of other strategies is averaged over 3 random seeds with standard error shown as confidence interval. }
    \label{tab:kw sensitivity}
    \vspace{-\intextsep}
\end{table*}
\subsection{Evaluation by GPT-4} \label{sec:alpaca}
To further demonstrate the annotation cost savings or various experimental design methods, we evaluate models finetuned on 45K prompts against the model trained on 90K prompts using random sampling with GPT-4 turbo as the judge, which displays a high agreement rate with ground truth human annotations . We follow the AlpacaEval framework~\citep{alpaca_eval} and report the win rate against the model finetuned on 90K randomly selected prompts. The models are evaluated on the 805 prompts from the standard AlpacaEval set~\citep{alpaca_eval}.
As showin in Figure~\ref{fig:alpaca}, multiple strategies reliably achieve the same or better performance while only taking 50\% of the annotation budget as compared to the model trained on 90K random prompts.  Compared to result from standard benchmark, we again observed that Facility Location ($\gamma$ = 0.002) strategy gains dominating improvement. On the other hand, margin-based methods also exhibited significant improvements, while other uncertainty-based methods performed similarly or even worse than random sampling. This suggests that, although uncertainty-based methods may still be helpful in aligning with human preference, selecting the right metric is crucial. Overall, compared to standard benchmark results, the evaluation by GPT-4 turbo highlights even greater potential for experimental design in instruction finetuning. However, further investigation into how different strategies relate to various metrics (i.e., the model's ability to perform different tasks) is needed to better understand the differences between standard and AlpacaEval benchmark results.

\subsection{Hyperparameter Selection for Facility Location} \label{sec:hyperparam}

In this section, we describe how to choose hyperparameter $\gamma$ before annotation begins. For a submodular function to be able to generate good summaries upon maximization, it should not \emph{saturate}. That is, the facility location objective (equation~\eqref{eq:facility_location}) eventually increases only minimally by increasing the budget $k$, which suggests the new elements no longer remain representative of the downstream task. Formally, let $S_1 \subset S_2 \subset  \ldots \subset S_k$ denote the greedy solution up to size $k$ and $F(S_k) = \sum_{i \in X} \max_{j \in S_k} w_{ij}$ denote the facility location function evaluation for set $S_k$ (defined in equation~\eqref{eq:facility_location}). The \emph{gain} by adding the $k$-th element is defined as $ F(S_{k}|S_{k-1}) \triangleq F(S_k) - F(S_{k-1})$. In the case of RBF kernel, kernel width $\gamma$ controls the degree of saturation. Higher values of $\gamma$ lead to decreasing gain by adding additional elements. On the other hand, very low values such as $10^{-4}$ will result in the kernel becoming close to the diagonal matrix, and therefore not being able to capture the interactions between data points. 


\begin{figure}[h!]
    \centering
    \includegraphics[width=\linewidth]{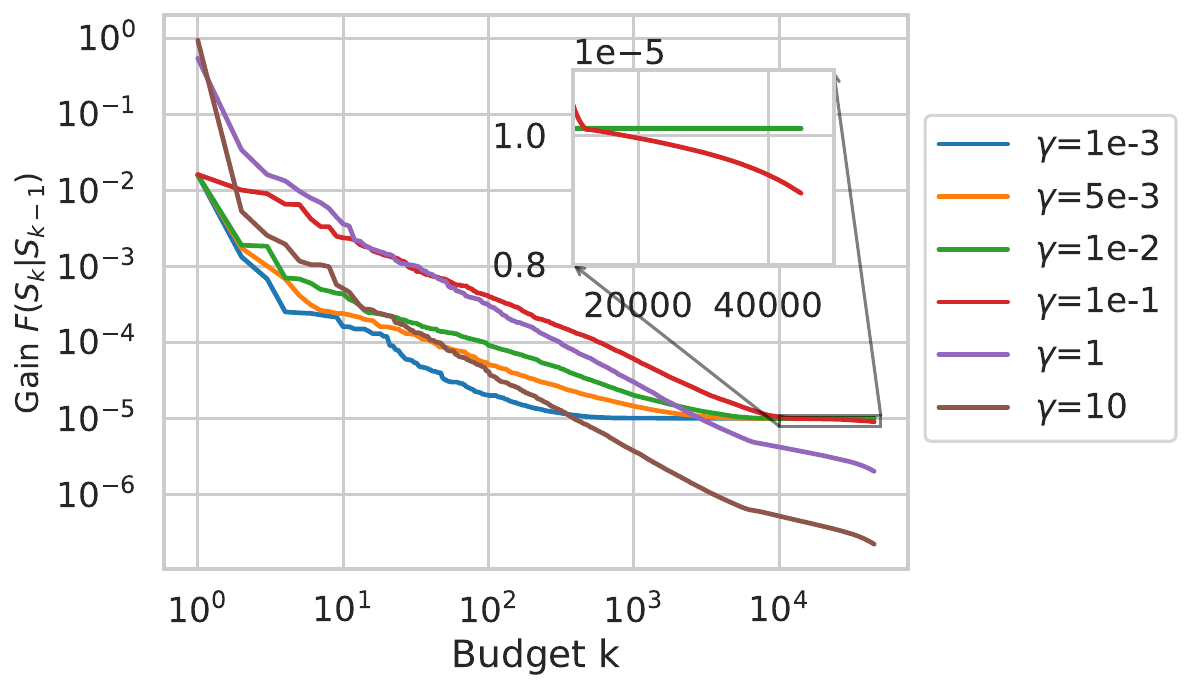}
    \caption{Plot of gains with set size as the course of greedy maximization for different kernel width $\gamma$; we run the greedy procedure till the budget of 45K is reached. The observed trend reveals that for higher $\gamma$, gains tend to attain a very small value (and continue to decrease linearly) even before 1K (for $\gamma = 10$) and 10K (for $\gamma = 1$) elements are selected. Although reducing $\gamma$ helps, gains continue to decrease sublinearly ($\gamma = 0.1$ after 20K). Notably, gains exhibit relative stability for $\gamma \in \{10^{-3}, 5\times 10^{-3}, 10^{-2}\}$ until we reach the desired budget of 45K, suggesting a potential range for $\gamma$.}
    \label{fig:gains}
    \vspace{-\intextsep}
\end{figure}

Therefore, to determine the potential range of kernel width, in Figure~\ref{fig:gains} we first visualize the gains of the submodular function when selecting 45K examples, for $\gamma \in \{10^{-3}, 5\times 10^{-3}, 10^{-2}, 10^{-1}, 1, 10\}$. The observed trend reveals that for higher $\gamma$, gains tend to attain a very small value (and continue to decrease) even before 1K (for $\gamma = 10$) and 10K (for $\gamma = 1$) elements are selected. Although reducing $\gamma$ helps, gains continue to decrease sublinearly ($\gamma = 0.1$ after 20K). Notably, gains exhibit relative stability for $\gamma \in \{10^{-3}, 5\times 10^{-3}, 10^{-2}\}$ until we reach the desired budget of 45K, suggesting that we can safely choose $\gamma$ in between these values.



With the range of potential $\gamma$ determined, we run an ablation study on 45K budget with four different kernel widths within this range to demonstrate the robustness of any hyperparameter in this range. As detailed in Table~\ref{tab:kw sensitivity}, the performance is consistently better than random selection strategy across different $\gamma$, which suggests the Facility location methods are less subject to hyperparameter changes once the appropriate range for $\gamma$ has been identified.

\section{Computational Complexity}
\label{sec:compute}

We primarily used A100 and A40 GPUs. We pre-compute the embeddings which takes 12 hours. For A100, each trial takes roughly 24 GPU hours including the evaluation for 90K budget, 15 GPU hours for 45K and 9 hours for 30K respectively. For A40, roughly 30 GPU hours for 90K budget.  

\section{Conclusion and Future Work}
Overall, we proposed using experimental design for finetuning LLMs with fewer annotated instructions. We present several heuristics that can be used to determine which set of instructions to annotate, and for the first time, demonstrate significant gains in label efficiency on generative tasks. 
Future research directions may include the following: 
\begin{itemize}
    \item Our work demonstrates the first empirical results of experimental design for SFT, but future work can focus on devising new methods within this framework to further improve label efficiency.
    \item Our setting finetunes the LLM only on the annotated instructions; future work could improve the utilization of unlabeled samples. 
\end{itemize}


\section*{Limitations}
While we successfully demonstrate the efficacy of experimental design on SFT, there are a few limitations of this work that we discuss. Firstly, we focus our experiments on the 7B version of Llama-2 since it is the most common and requires the least compute. Secondly, we conduct our experiments on curated datasets where we assume accurate responses to prompts are available. However, we do not consider the effect of inaccurate/noisy responses which can be prevalent when procuring responses from crowd-workers. Finally, we evaluate our model after SFT, and do not consider the effect of reinforcement learning with human feedback \cite{ouyang2022training} which is frequently applied after SFT.  

\section*{Ethics Statement}
Our work proposes a computationally inexpensive method of reducing the number of labeled instructions needed for SFT while preserving the performance of the resulting LLM. In doing so, we also mitigate the carbon footprint associated with training such models, which is critical given growing concerns about the environmental impact of modern ML systems. Furthermore, researchers and organizations with limited compute and insufficient financial resources to procure a vast number of crowd-workers can benefit from our work. 

It is important to note, however, that all of the experimental design techniques that we propose are forms of \emph{biased} sampling. If the subset of data chosen for annotation is not representative, it is possible biases of pretrained model will be amplified upon finetuning. Thus, it is critical for practitioners to ensure that the subsets selected by our experimental design framework are inclusive. 


\section*{Acknowledgements}
JZ and RN would like to thank the support of AFOSR/AFRL grant FA9550-18-1-0166, the UW-Madison Data Science Institute and the NSF grants DMS-2023239, CNS-2112471. GB, AD, and JB would like to thank NSF Grant Nos.\ IIS-2106937 and IIS-2148367, and by NIH/NHGRI U01 award HG009395.  
\newpage
\bibliography{anthology,custom}
\bibliographystyle{acl_natbib}

\newpage
\end{document}